# Image Segmentation Based on the Self-Balancing Mechanism in Virtual 3D Elastic Mesh

XIAODONG ZHUANG[1,2], N. E. MASTORAKIS[2], JIERU CHI[1], HANPING WANG[1]

1. Qingdao University, Automation Engineering College, Qingdao, 266071 CHINA
2. Technical University of Sofia, Industrial Engineering Department,
Kliment Ohridski 8, Sofia, 1000 BULGARIA

*Abstract:* -In this paper, a novel model of 3D elastic mesh is presented for image segmentation. The model is inspired by stress and strain in physical elastic objects, while the repulsive force and elastic force in the model are defined slightly different from the physical force to suit the segmentation problem well. The self-balancing mechanism in the model guarantees the stability of the method in segmentation. The shape of the elastic mesh at balance state is used for region segmentation, in which the sign distribution of the points' $z$ coordinate values is taken as the basis for segmentation. The effectiveness of the proposed method is proved by analysis and experimental results for both test images and real world images.

*Key-Words:* - Image segmentation, elastic mesh, self balancing, physics inspired method, stress and strain

## 1 Introduction

Self balancing is a mechanism exists in many natural systems. In such phenomena, a system has the mechanism of counteracting the external influence to weaken the change brought by those external interventions. For example, when a solid elastic object deforms due to external force, there will be internal force (stress) between the local parts inside the object emerging simultaneously with the deformation [1-4]. The stress resists the effect of the external force, which may prevent further more deformation if they reach equilibrium [1-4]. A simple demonstration is shown in Fig. 1.

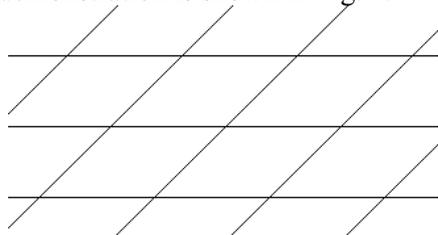
(a) original shape of the elastic grid without any external force

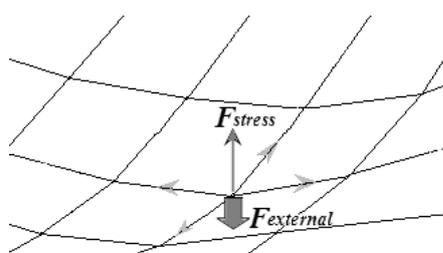
(b) the deformation of elastic grid caused by the external force, and the effect of stress

Fig. 1 A simple demonstration of stress on an elastic grid

Another example can be found in chemical reversible reaction [5]. When increasing the amount (or density) of one reactant, the reaction will be promoted to the direction of consuming more reactant whose amount has just been increased until new balance is reached, which is also the effect of weaken the change brought by external influence with balancing mechanism. Similar phenomena can be found in other natural systems such as gas or solution diffusion, balance of food chain, etc.

In the self balancing mechanism mentioned above, the system's state at new balancing point depends on the external influence (such as the number and amplitude of external forces, the amount of reactant added, etc.). In another word, the internal change (such as the deformation of an elastic object under forces) reflects the feature of the external influence. This idea may be the inspiration (or starting point) of designing novel methods for problem solving, if the self balancing mechanism matches the nature of the problem well.

Image segmentation is a basic and important problem in image processing, which is fundamental for many practical tasks [6-8]. There have been extensive studies about image segmentation, and it is still one research focus in image processing. Many research efforts focus on this topic each year, and different methods have been invented for different practical tasks [6-12]. Practically, different segmentation methods suit different image types respectively. Image segmentation has been mapped to various solution frameworks such as classification, optimization, or nature-inspired methods [6,11-18].

In the authors' previous work, a virtual elastic grid limited on 2D image plane was proposed for image segmentation, which limited the grid nodes only to move on the 2D image plane [12]. In order to obtain better segmentation results, and also to extend method's adaptability for different image types, a novel method is presented in this paper, which can be regarded as the reformation of the previous method in [12]. A novel 3D elastic mesh is presented in the new method imitating the physical mechanism of stress and strain, which allows the nodes in the mesh to move in 3D space. In another word, the new method maps the segmentation problem to the deformation of a virtual 3D elastic mesh with analogy to the physical stress and strain in elastic objects.

## 2 The Virtual Elastic Mesh on Digital Image

Consider the problem of judging whether two adjacent image points belong to the same region. There are two aspects to be considered. First, if the features of the two pixels (such as greyscale, colour, etc.) are quite different, it is more likely that they belong to different regions. The more the difference between two adjacent pixels, the more likely they should be separated from each other in segmentation. In this paper, for an adjacent pixel pair, a virtual force is defined to represent the difference between adjacent pixels, which is named the "repulsion force":

$$F_r = K_1 \cdot (g - g_a) \qquad (1)$$

where $F_r$ is the virtual repulsive force applied on a point by its adjacent neighbour point; $g$ and $g_a$ are the greyscale values of the point and its neighbour respectively; $K_1$ is a positive coefficient predefined. Here the positive direction of force is along the $z$ coordinate axis in space. And the greyscale is taken as the feature for segmentation. The more greyscale difference, the larger the repulsion force between adjacent pixels. Moreover, if $g$ is larger than $g_a$ (i.e. the point has larger greyscale value than its adjacent neighbour), it is pushed upward by its neighbour (i.e. its $z$ coordinate will increase); otherwise, it will be pushed downward.

Secondly, regions are always local parts in images, which indicates that pixels close to each other may belong to the same region. Usually, this assumption is reasonable for most part of the image, except the region borders. Therefore, another virtual force is defined to represent the possible connection between adjacent pixels, which is named the "elastic force":

$$F_s = K_2 \cdot (z_a - z) \qquad (2)$$

where $F_s$ is the virtual elastic force (the force of stress) applied on a point by its adjacent neighbour; $z$ and $z_a$ are the $z$ coordinate values of the point and its adjacent neighbour respectively; $K_2$ is a predefined positive coefficient. If there is displacement between two adjacent pixels with respect to their original positions, the elastic force has the effect of pulling them closer. The larger the displacement, the larger the elastic force. Moreover, if $z$ is larger than $z_a$ (i.e. the point is lower than its neighbour on $z$ coordinate), it will be pulled upward by its neighbour; otherwise it will be pulled down. In either case, the difference of $z$ coordinate will be decreased by the elastic force. It should be noted again that for both the virtual repulsive force and the elastic force, the force direction is along the $z$ coordinate axis in space for simplicity in simulation. A simple demonstration of the repulsive and elastic force for two adjacent pixels in image is shown in Fig. 2, where the pixels are represented by squares with corresponding greyscale value, and their heights represent the values of $z$ coordinates. In Fig. 2, $A$ and $B$ are two adjacent pixels, and $A$ has larger greyscale value than $B$. There is repulsive force between $A$ and $B$, which is represented by $Fr_{AB}$ and $Fr_{BA}$. And there is also elastic force, which is represented by $Fs_{AB}$ and $Fs_{BA}$. In another word, the forces on $A$ applied by $B$ are $Fr_{BA}$ and $Fs_{BA}$ respectively.

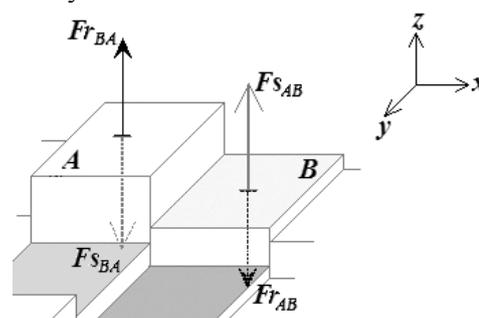

Fig. 2 A demonstration of the repulsive and elastic force in digital image

Based on the above two virtual forces, a virtual elastic mesh for image segmentation can be established based on an analogy of physical strain and stress. All the image points are considered as the points in the mesh, and there are elastic connections between adjacent image points. Initially, all the points are on the same image plane. The space position of each point is described by three coordinates: $x$, $y$ and $z$, where $x$ and $y$ are the point's positions on the image plane. For each point, it applies two kinds of virtual force on its adjacent points: the "repulsion force" and the "elastic force", which will cause the points to move. For simplicity, it is supposed that all the points can only move vertically (i.e. perpendicular to the image plane). Therefore, in the dynamic deformation of the mesh shape, the *x* and *y* coordinates of each point keep unchanged, and only the *z* coordinate can change by the virtual force.

## 3 Image Segmentation Based on the Virtual Elastic Mesh

### 3.1 The simulation process

The two kinds of virtual forces between adjacent points are analyzed as follows. For any pair of adjacent points *A* and *B*, if *A* has larger greyscale than *B*, *A* applies a virtual repulsive force to push *B* down; otherwise, *A* applies a repulsive force to push *B* up. The larger the greyscale difference, the larger the repulsive force. In either case, the repulsive force causes the adjacent points to get away. On the other hand, for any pair of adjacent points *A* and *B*, if *A* has a larger value of *z* coordinate than *B*, *A* applies a virtual elastic force to pull *B* up; otherwise, *A* applies an elastic force to pull *B* down. In either case, the elastic force causes the adjacent points to get closer. The opposite effects of the repulsive and elastic force may reach balance for an image after a dynamic process, and the shape (i.e. the *z* coordinates of the image points) of the virtual mesh may become stable at balance, which may reveal some feature of image structure.

There is a difference between the repulsive force and the elastic force: the repulsive force depends totally on the greyscale difference between adjacent pixels, and keeps unchanged for a pair of adjacent points; while the elastic force depends on the height difference (i.e. the difference between the *z* coordinates) of adjacent points, which is changing in the process.

Since the method is an artificial simulation imitating the physical strain and tress in elastic materials, there are several simplifications in the above process compared to the actual physical system. First, the movement of points is limited to the vertical direction, i.e. only the *z* coordinate is changed in the process. Second, the virtual elastic force is defined as proportional to the difference of *z* coordinates between adjacent points, not the three-dimensional distance between them. Third, the change of a point's *z* coordinate is defined as proportional to the total force on it:

$$\Delta z = K_3 \cdot F_{net} \qquad (3)$$

where $\Delta z$ is the variation of the point's *z* coordinate due to the resultant force $F_{net}$ by summing up all the repulsive and elastic forces from its adjacent points; $F_{net}$ is the resultant force considering all the adjacent points; $K_3$ is a predefined positive coefficient. However, in physics the total force only determines the change of velocity. Because the model is an artificial one which exploits the physical mechanism of self balancing, the above simplification is reasonable for simulation, and the effectiveness can be proved by the experimental results shown in following sections.

The steps of the simulation process are as follows:

*Step1* Initialize the *z* coordinates of all the points as zero;
*Step2* Calculate the repulsive force for each pair of adjacent points according to their greyscale difference;
*Step3* Calculate the elastic force for each pair of adjacent points according to their *z* coordinate difference;
*Step4* For each image point, calculate the net force $F_{net}$ by summing up all the repulsive and elastic forces from its adjacent points (4-connection adjacent points);
*Step5* For each point, update its *z* coordinate with the calculated $\Delta z$;
*Step6* If the mesh is close enough to the balance state, stop the simulation; otherwise, return to *step3*. The stop condition can be implemented by comparing the average change of *z* to a predefined small threshold.

### 3.2 Model analysis and simulation results for simple test images

A detailed analysis of the above process is as follows. At the beginning, all the points are on the same image plane, and there is only repulsive force but no elastic force between adjacent points. The repulsive force causes the adjacent points with greyscale difference to separate in *z* coordinate. The vertical height difference in turn causes the emergence of elastic force, and the deforming of the mesh shape begins. The virtual elastic mesh deforms its shape until a balance state is reached.

The effect of the above dynamic process can be analyzed for two different cases. The first case is for adjacent pixel pairs of large greyscale difference, such as the points at the region borders. In such case, the repulsive force is strong because of the large greyscale difference. The relatively large repulsive force can causes large difference in *z* coordinate between adjacent pixels, and the height difference in turn causes the elastic force to counteract the mesh deformation. In this case, the strong repulsive force is the dominant factor. As a consequence, for the pixel pair, the z coordinate of one pixel will become positive while the other will become negative. In another word, the pixel with higher greyscale will be pushed up above the image plane, while the one with lower greyscale will be pushed down below the image plane. This just represents the differentiation of two different regions.

The second case is for adjacent pixel pairs with small greyscale difference, such as adjacent pixels within the same region. Here the repulsive force is weak due to the small greyscale difference. Therefore, the dominant factor in this case is the elastic force, especially when there is obvious height difference between adjacent points. The elastic force tends to decrease the height difference between adjacent points. It counteracts the deformation of the mesh, and tends to draw the adjacent points back to the same height. As a consequence, the points within a same region tend to have the same sign of $z$ coordinate (i.e. all above or all below the image plane within the same region).

A brief demonstration of the above process is shown in Fig. 3. The arrows in Fig. 3 represent the repulsive force or elastic force defined in Section 2. As shown in Fig. 3, the eight pixels (or nodes) on the left side have higher greyscale value than those eight pixels on the right side, therefore there is greyscale difference between the two sides. Fig. 3(a) shows the initial state of the virtual mesh, in which all the points in the image have the same value of $z$ coordinate (i.e. the same height), and the only force in the mesh is the repulsive force at region borders (i.e. no elastic force at beginning). It can be seen in Fig. 3(a) that all the nodes of the mesh are on the same plane (i.e. $x$-$y$ plane or image plane). In Fig. 3(b), the height difference occurs across the region boundary due to the repulsive force, which in turn causes elastic forces between adjacent points. Fig. 3(c) shows the ideal final state of the virtual mesh at balance, in which the repulsive and elastic forces are at balance at region borders, and the heights of the points become identical (i.e. no force exists) inside each homogeneous region of uniform greyscale. It can be seen in Fig. 3(c) that the nodes on the right have been lifted above their original position on $z$ direction, while the nodes on the right have been pressed down below their original position on $z$ direction. The mesh shape at balance state can provide clues for segmentation, because the adjacent regions have opposite sign of $z$ coordinate (i.e. opposite sign of height value).

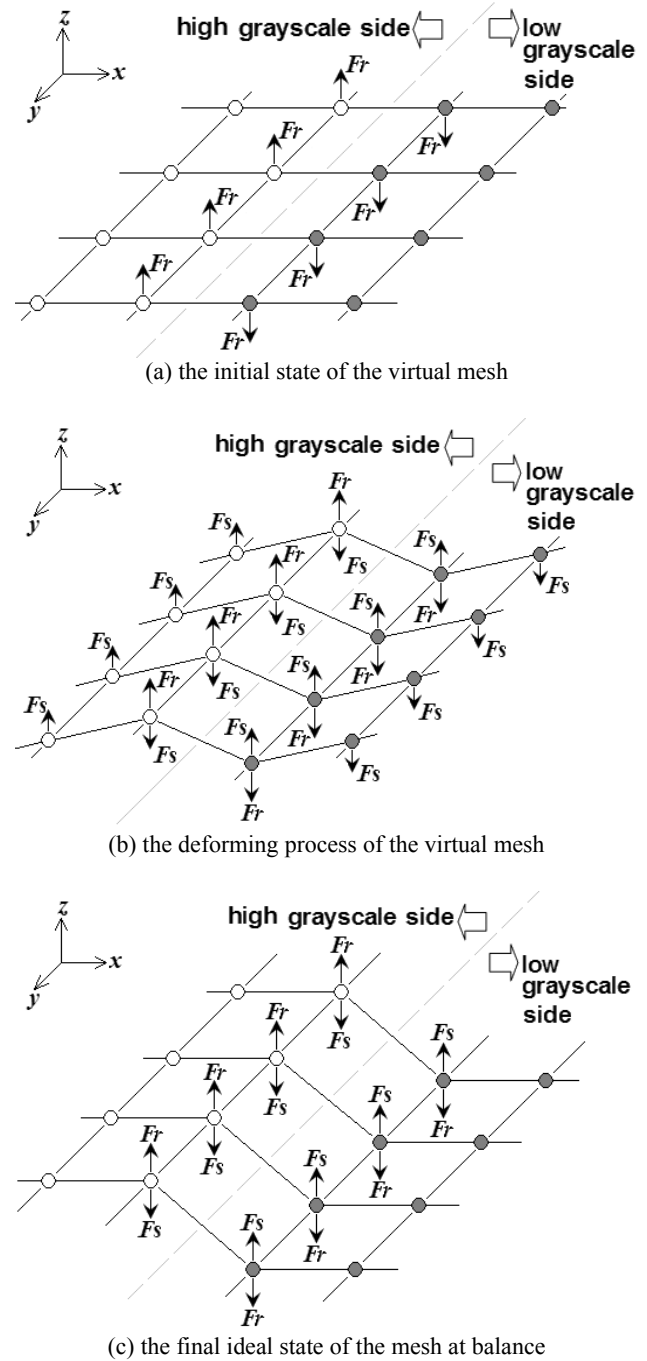

(a) the initial state of the virtual mesh

(b) the deforming process of the virtual mesh

(c) the final ideal state of the mesh at balance

Fig. 3 Demonstration of the evolving process of the virtual elastic mesh on digital computer

The difference between the above two cases just satisfies the requirement of differentiating adjacent regions. At the boundary of two different regions, the points' $z$ coordinate difference is large, and the sign of the $z$ coordinate reverses across the boundary after mesh deforming, because at one side the points are pushed up above the image plane, while at the other side the points are pushed down below the plane. However, the $z$ coordinate of points within a same region has a tendency of uniform distribution by the effect of elastic force. The final result of mesh deforming may reflect the differentiation of adjacent regions, which will provide clues for segmentation.

The simulation experiments have been carried out for a group of simple but typical test images by programming in C. Some of the results are shown in Fig. 4 to Fig. 9.

Fig. 4 shows the simulation results of the mostly simple case: the left and right half of the image are of different greyscale, which can specifically reveal the effect of the model on region borders. In order to demonstrate the process of mesh deforming step by step, the intermediate results at several specific simulation step numbers are recorded. In Fig. 4(b) to Fig. 4(f), the white and black points represent the positive and negative sign of height (i.e. $z$ coordinate) respectively, and the gray points represent that the height is zero (i.e. not a definite sign yet). In Fig. 4(b), it can be seen that shortly after the simulation starts, only the points close to the region border have a definite sign of height. From Fig. 4(c) to Fig. 4(f), it is clear that the part of definite sign of height expands with the increasing of simulation time. Finally, in Fig. 4(f) each point has a definite sign (positive or negative) of height, and the final result of sign distribution just corresponds to the segmentation of the two halves of the image by different sign of height.

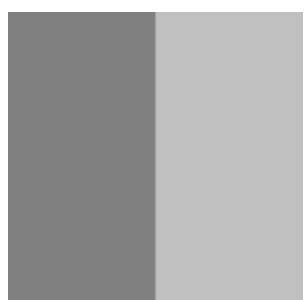

(a) Test image 1

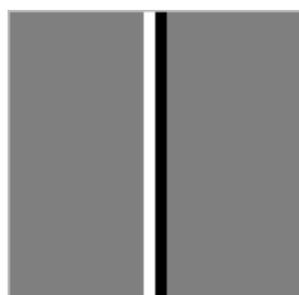

(b) sign distribution of net carrier after 5 simulation iterations

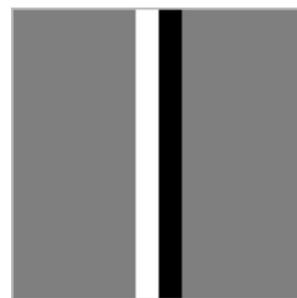

(c) sign distribution of net carrier after 10 simulation iterations

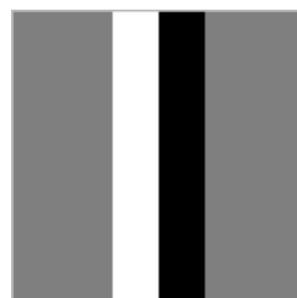

(d) sign distribution of net carrier after 20 simulation iterations

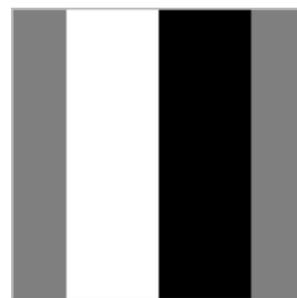

(e) sign distribution of net carrier after 40 simulation iterations

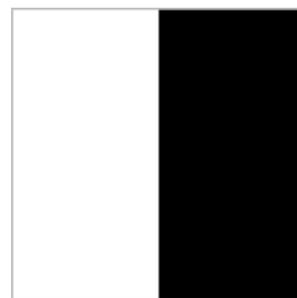

(f) sign distribution of net carrier after 80 simulation iterations

Fig. 4 The experimental results for Test image 1

Fig. 5(a) shows another test image, which has a rectangle region. The intermediate results are shown from Fig. 5(b) to Fig. 5(f). It is clear that the positive sign part expands outward from the rectangle border, while the negative sign part expands inward with the increasing of simulation time. Similarly, the final result of sign distribution of heights in Fig. 5(g) can also support a definite segmentation of the Test image 2.

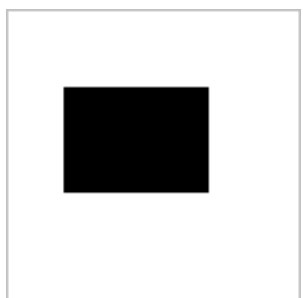

(a) Test image 2

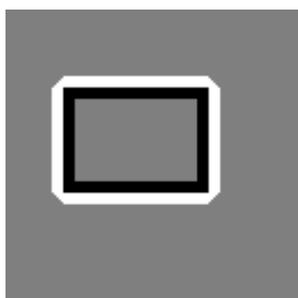

(b) sign distribution of net carrier after 5 simulation iterations

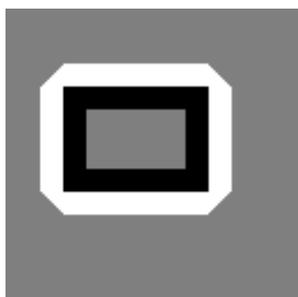

(c) sign distribution of net carrier after 10 simulation iterations

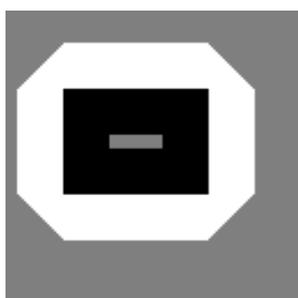

(d) sign distribution of net carrier after 20 simulation iterations

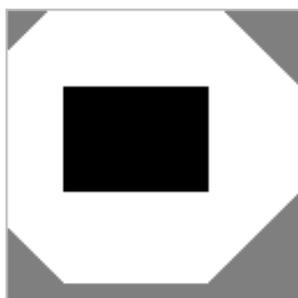

(e) sign distribution of net carrier after 40 simulation iterations

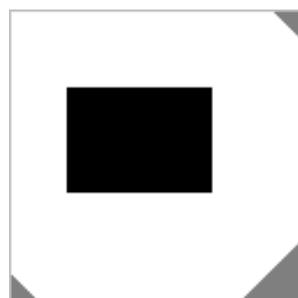

(f) sign distribution of net carrier after 60 simulation iterations

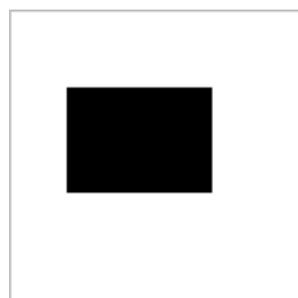

(g) sign distribution of net carrier after 100 simulation iterations

Fig. 5 The experimental results for Test image 2

In order to give a clear and visible impression of the deforming elastic mesh, the height (i.e. $z$ coordinate) of each point is recorded in the experiment. Fig. 6 shows mesh shape in 3D view for Fig. 5(g) after 100 simulation iterations. The $x$ and $y$ coordinate in Fig. 6 are the coordinates of points on the image plane (i.e. the $x$-$y$ plane with $z$=0 corresponds to the image plane). The $z$ coordinate represent the height of the mesh nodes. Since the rectangle region in Fig. 5(a) has lower greyscale than the background region, it can be clearly seen that the rectangle region corresponds to a valley of low height in Fig. 6. The points within that valley have negative sign of $z$ coordinate, while the points outside it have positive height values. That can provide the basis for the segmentation.

Fig. 7(a) shows a test image which has three regions of different shape and greyscale. The intermediate results are shown from Fig. 7(b) to Fig. 7(e). The process of region formation by the mesh deformation step by step can be clearly seen. In Fig. 7(f), the three regions in Test image 3 have negative sign of height value, but the background has positive height value.

In the experiment for Test image 3, the height (i.e. $z$ coordinate) of each point is recorded. Fig. 8 shows the mesh shape in 3D view for Fig. 7(f) after 80 simulation steps. The $x$ and $y$ coordinate in Fig. 8 are the coordinates of points on the image plane. The $z$ coordinate represent the height of points. It can be clearly seen that the regions of circle, triangle and rectangle correspond to three valleys of low height in Fig. 8. The points within the three valleys have negative sign of *z* coordinate, while the points outside them have positive height values.

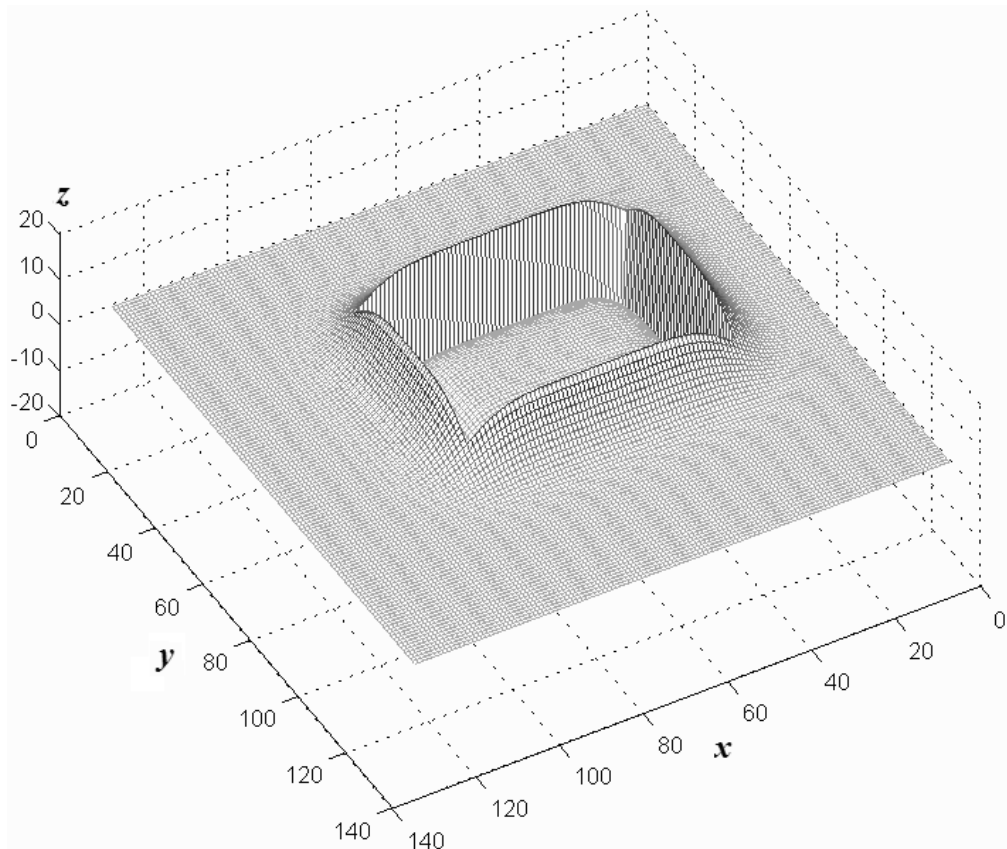

Fig. 6 The shape of the elastic mesh after 100 simulation iterations for Test image 2

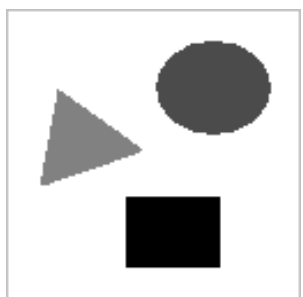

(a) Test image 3

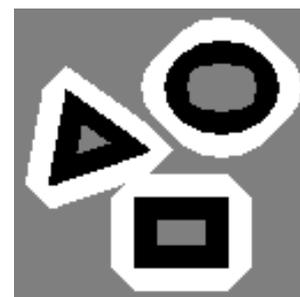

(c) sign distribution of net carrier after 10 simulation iterations

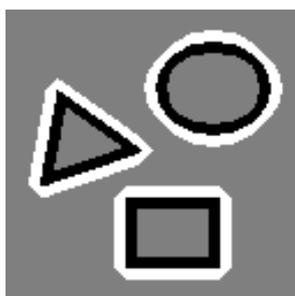

(b) sign distribution of net carrier after 5 simulation iterations

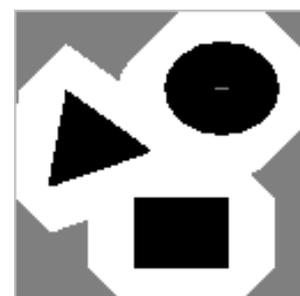

(d) sign distribution of net carrier after 20 simulation iterations

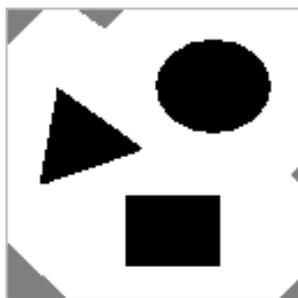

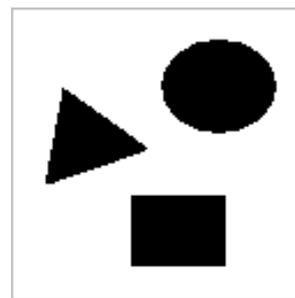

(e) sign distribution of net carrier after 40 simulation iterations

(f) sign distribution of net carrier after 80 simulation iterations

Fig. 7 The experimental results for Test image 3

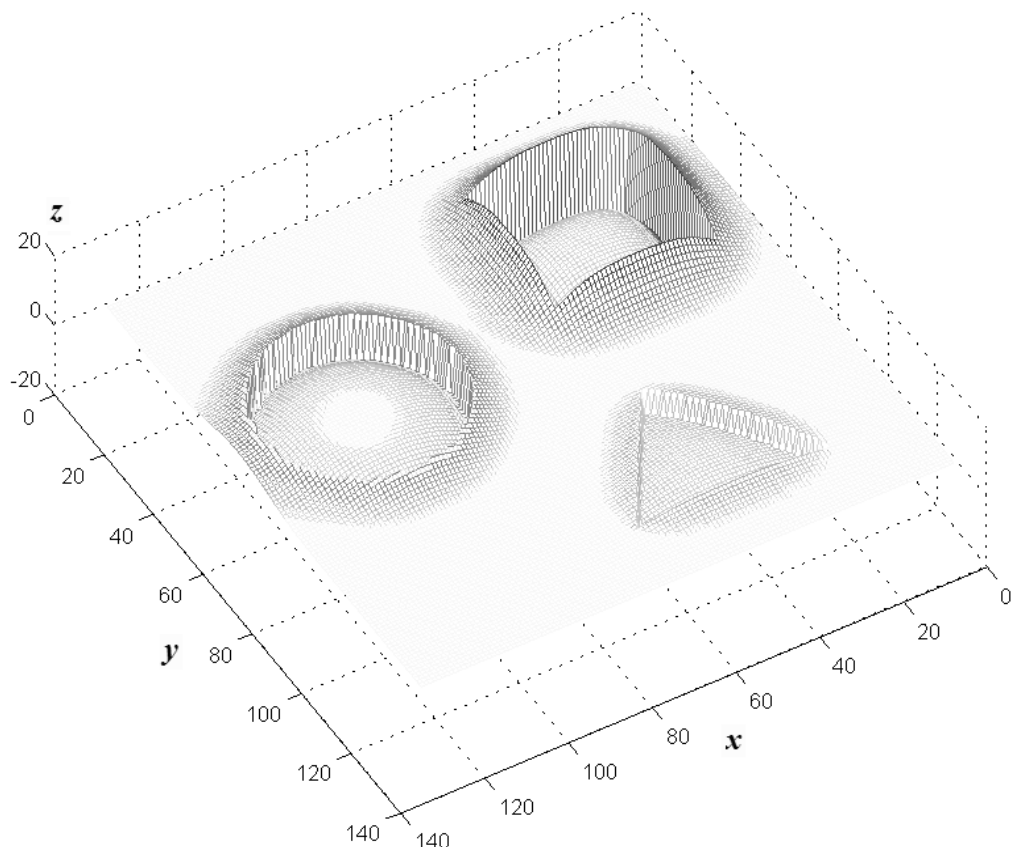

Fig. 8 The shape of the elastic mesh after 80 simulation iterations for Test image 3

In order to quantitatively investigate the mesh deformation process, the average of the absolute value of $\Delta z$ for all the points is calculated and recorded as a measurement of mesh shape variation (i.e. the convergence degree to the balance state). Fig. 9 shows the relationship between that average value and the simulation time. In Fig. 9, the variation of mesh shape decreases with the increasing of simulation time, which indicates the mesh approaches the balance state with the simulation going on.

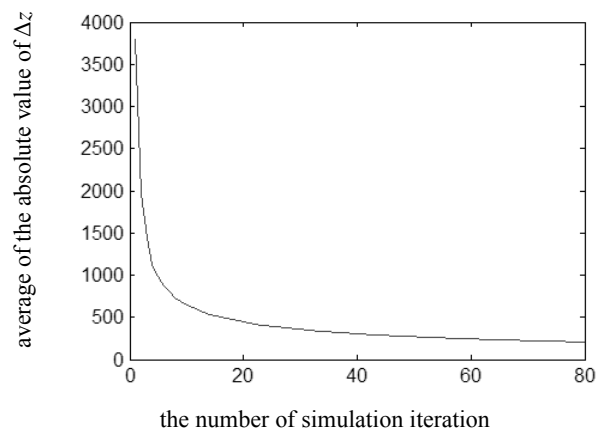

Fig. 9 The relationship between the average change of height and the simulation time (for Test image 3)

Here another feature of the proposed model is discussed as follows. Although the virtual repulsive force and elastic force are defined locally for each pair of adjacent points, the mesh is integrated as a whole by elastic connections between adjacent pixels. With the elastic connection, the interaction between local adjacent pixels can affect other image areas gradually by the evolving of the system. Finally, balance can be reached at both local and global level, and a certain shape of the mesh can emerge, which is useful to image segmentation.

### 3.3 Segmentation of real world images based on the virtual elastic mesh

In the above experimental results for the test images, it is shown that the sign of height values are opposite in two different adjacent regions, which can provide the basis of region separation. In order to obtain the segmentation result from the sign distribution of the height value, a region clustering approach is proposed as following:

*Step*1: Implement the simulation of elastic mesh deformation as proposed in section 3.1;

*Step*2: Obtain the sign distribution of the height value (i.e. *z* coordinate) in the mesh;

*Step*3: Group the adjacent points with the same sign of height value into connected regions. In the region clustering process, the adjacent pixels of the 4-connection (i.e. the upper, lower, left and right pixels) for an image point *p* is investigated. If any of the four adjacent pixels has the same sign of height value as *p*, it is grouped into the region which *p* belongs to. The obtained connected regions are the result of region segmentation.

The obtained set of connected regions is the result of region segmentation. Fig. 10 to Fig. 12 show the region clustering results for Fig. 4(f), Fig. 5(g) and Fig. 7(f), where different regions are represented by different gray-scale values.

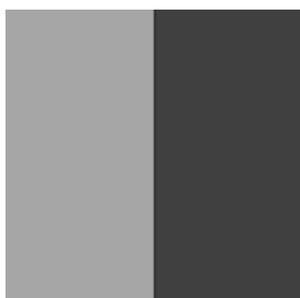

Fig. 10 The region clustering result for Fig. 4(f)

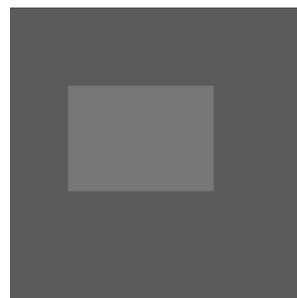

Fig. 11 The region clustering result for Fig. 5(g)

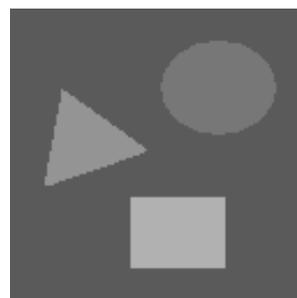

Fig. 12 The region clustering result for Fig. 7(f)

However, real world images are much more complex than the simple test images. To investigate the effect of the proposed method on real world images, experiments are carried out for a series of real world images. For demonstration, some of the results are shown in Fig. 13 to Fig. 18, which are for the broadcaster image, the peppers image, the locomotive image, the flower image and the medical heart image. The experimental results indicate that the proposed method can obtain large amount of regions (more than a hundred) because of the complexity of real world images. There are 169 regions obtained for the broadcaster image, 101 for the peppers image, 309 for the locomotive image, 943 for the flower image, and 400 for the medical heart image.

To obtain practically useful segmentation result, a region merging method is proposed for the above segmentation results of real world images based on the gray-scale similarity of adjacent regions. First, an expected number of remaining regions after merging is given (usually by trail). Then the following steps are carried out to merge regions until the expected region number is reached:

*Step*1: For each region in the image, calculate its average gray-scale value.

*Step*2: Find the pair of neighboring regions with the least difference of the average gray-scale, and merge them into one region.

*Step*3: If current region number is larger than the expected region number, return to *Step*1; otherwise, end the merging process.

In Fig. 13 to Fig. 17, the figures show the original image, the sign distribution of height value in the deformed mesh, the region segmentation results by clustering, and also the result of region merging by the method proposed above. In the sign distribution of height value, the white points represent positive $z$ coordinate, and black points represent the negative height value. In the region segmentation results and region merging results, different regions are represented by different greyscale values.

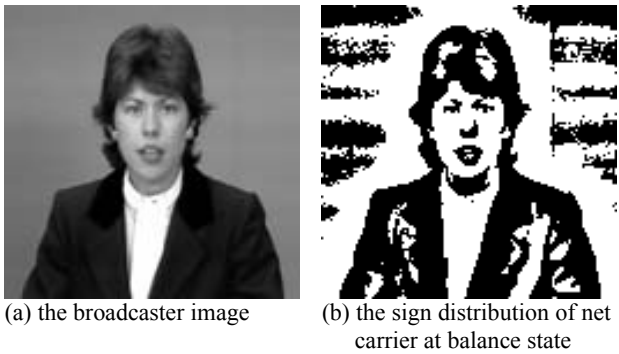

(a) the broadcaster image    (b) the sign distribution of net carrier at balance state

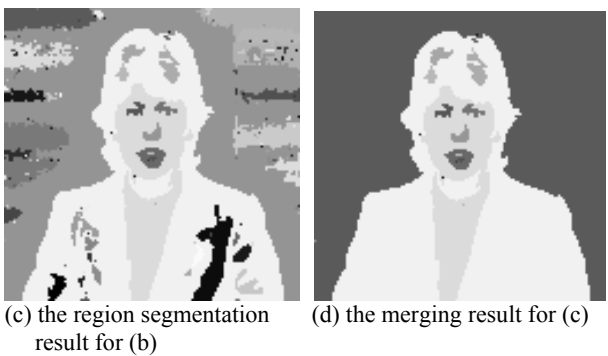

(c) the region segmentation result for (b)    (d) the merging result for (c)

Fig. 13 The experimental result for the broadcaster image

For the broadcaster image, the number of remained regions after merging is 20 in Fig. 13(d). Because the hair and the suit are both black, they are segmented as one region in Fig. 13(d). The eye and brow, nose and mouth are well segmented in the merged result.

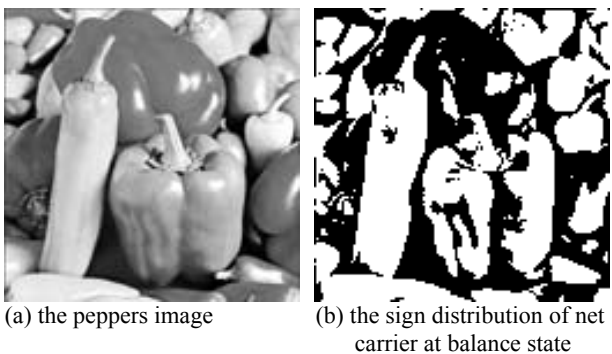

(a) the peppers image    (b) the sign distribution of net carrier at balance state

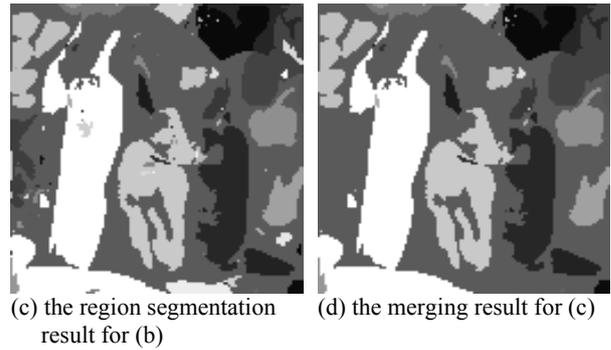

(c) the region segmentation result for (b)    (d) the merging result for (c)

Fig. 14 The experimental result for the peppers image

For the peppers image, the number of remained regions after merging is 50 in Fig. 14(d). Because the objects in this image pile up and are obscured, only some front objects are segmented completely in the merged results. It is reasonable to expect a better segmentation result for the colour version of the peppers image, because colour provides much richer information which may differentiate overlaid objects well.

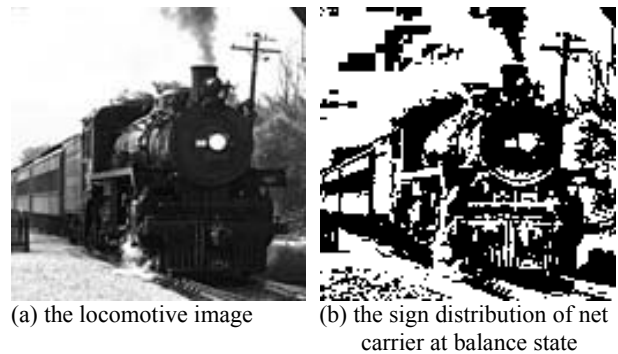

(a) the locomotive image    (b) the sign distribution of net carrier at balance state

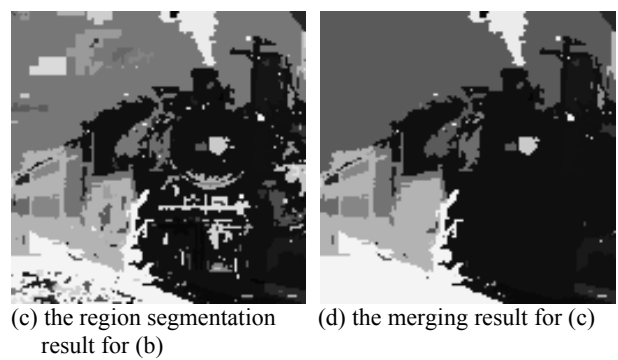

(c) the region segmentation result for (b)    (d) the merging result for (c)

Fig. 15 The experimental result for the locomotive image

For the locomotive image, the number of remained regions after merging is 100 in Fig. 15(d). The locomotive is well segmented from the background sky and land, but the background trees are not well separated from the locomotive. The separation of the locomotive and the tree can be a typical case to be further studied to improve the effectiveness of the proposed method.

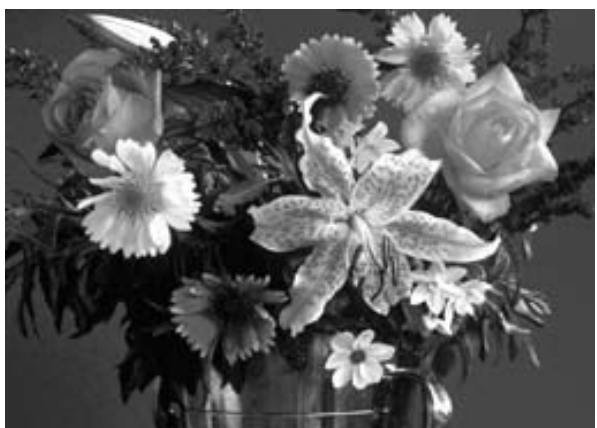

(a)   the flower image

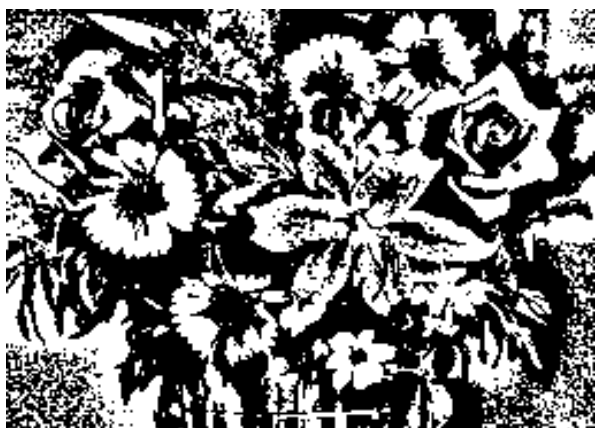

(b) the sign distribution of net carrier at balance state

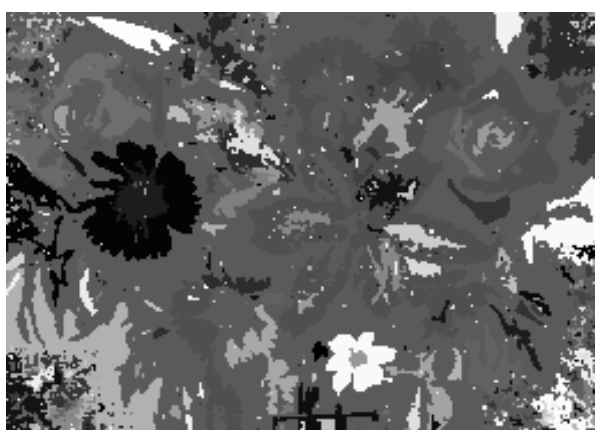

(c) the region segmentation result for (b)

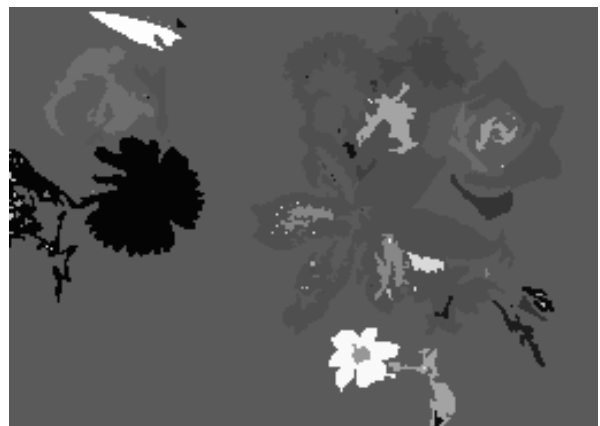

(d) the merging result for (c)

Fig. 16 The experimental result for the flower image

For the flower image, the number of remained regions after merging is 100 in Fig. 16(d). Because the leaves have similar low greyscale as the dark background, only the front white flowers are well segmented in the merged results. Like the peppers image, it is also reasonable to expect a better segmentation result for the colour version of this image.

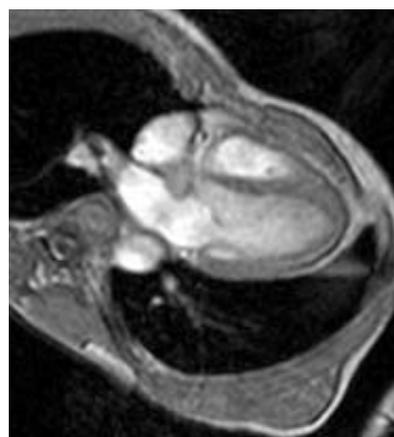

(a) the flower image

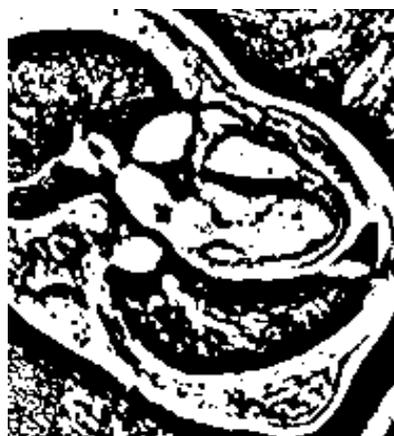

(b) the sign distribution of net carrier at balance state

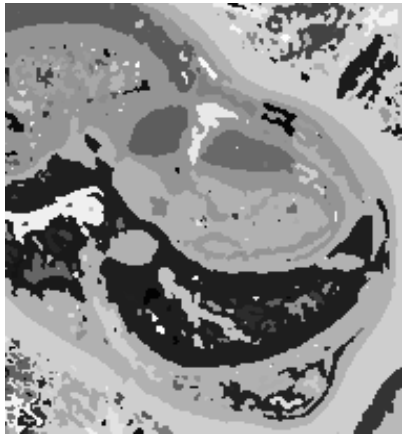

(c) the region segmentation result for (b)

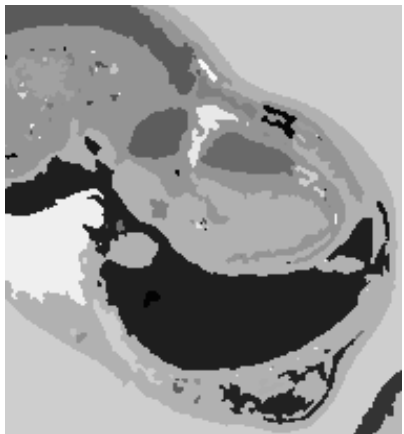

(d) the merging result for (c)

Fig. 17 The experimental result for the medical heart image

In Fig. 17(d), the number of remained regions after merging is 100. Fig. 17(d) clearly shows the heart structure. Moreover, in each iteration of simulation, the average of the absolute value of $\Delta z$ for all the points is calculated and recorded as a measurement of the convergence degree to the mesh's balance state, which is shown in Fig. 18. Fig. 18 indicates that the mesh approaches the balance state with the simulation going on.

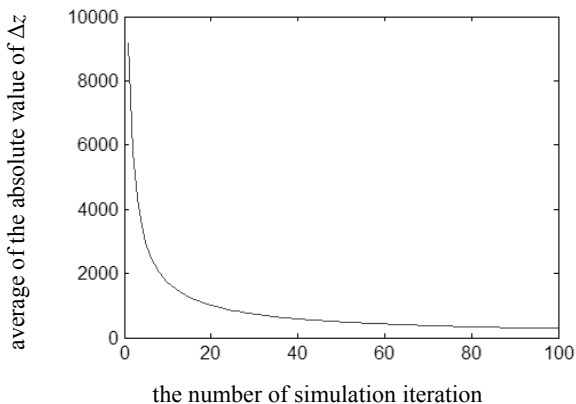

Fig. 18 The relationship between the average change of height and the simulation time (for the medical heart image)

In implementation of the above region-merging process, a 'flag matrix of adjacency' was used as an efficient representation of relationship between regions, where each matrix element $a_{ij}$ represents whether the $i$-th region and the $j$-th region are adjacent or not (e.g. $a_{ij}$ =1 means they are adjacent, while $a_{ij}$ =0 means they are not). The optimization in algorithm implementation guarantees the computational efficiency. In the experiments the method shows nearly real-time performance by programming with VC++ on a platform of Intel Core i3 3.6G CPU.

The above experimental results indicate that the proposed method is effective in segmentation of real world images. In the experiments, relatively large amount of regions can be obtained by grouping (or clustering) the sign of point's height due to the complexity of real world images. From the results, it can be seen that main object regions can be well segmented, and some regions are segmented in good detail. However, sometimes, at some part of object borders, two objects are not well separated due to reasons like greyscale similarity. For such cases, the greyscale feature alone may not be sufficient for accurate segmentation. It is indicated that other image features such as colour and texture may be introduced into segmentation for improvements.

Moreover, some comparison is shown in Fig. 19 between the new method in this paper and the previous method in [12]. The segmentation results for Fig. 13(a) and Fig. 14(a) are shown as examples for comparison. Fig. 19(a) and (c) are the results of previous method in [12], while Fig. 19(b) and (d) are the results of the new method in this paper.

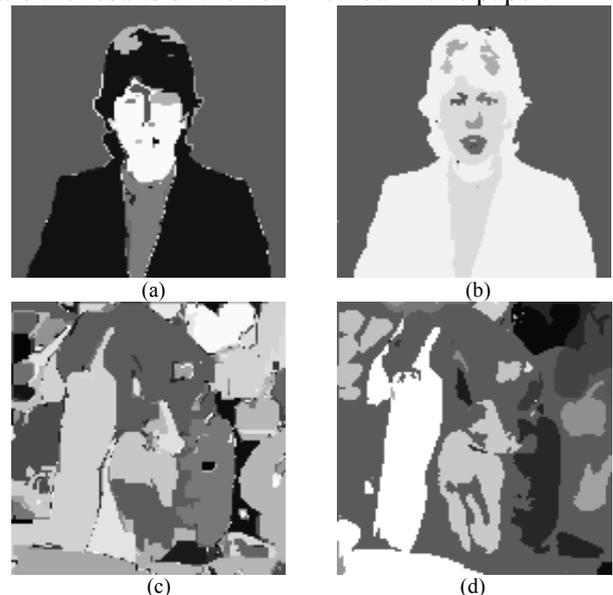

Image (a) and (c) are results of previous method in [12], while image (b) and (d) are results of new method in this paper

Fig. 19 The comparison between the segmentation results of the new method and the previous on in reference [12]

Although currently the evaluation of segmentation results often alters for different types of images, some preliminary evaluation can still be done by subject perception on the above segmentation results. By comparison of the results in Fig. 19, the new method proposed in this paper produced more accurate details of segmentation (such as the regions of eyes and mouth in the broadcaster image).

In order to investigate the intrinsic feature of the proposed method, the comparison has also been done with another well-known nature-inspired method – the watershed algorithm for image segmentation. The watershed algorithm takes the image as natural terrain or landform with hills and valleys, and it segments the image by imitating the rainfall or water-logging process to find the watershed lines between the water pools. Besides their common feature of nature-imitating, the two methods can both segment images into large amounts of preliminary regions without some pre- or post-processing steps. In order to compare their intrinsic feature, experiments were done to obtain the segmentation results of the two methods without any pre- or post-processing. Example results are shown in Fig. 20, where the preliminary segmentation results for Fig. 13(a) and Fig. 14(a) are shown as (a),(b) and (c)(d) respectively. The preliminary regions are shown by their borders.

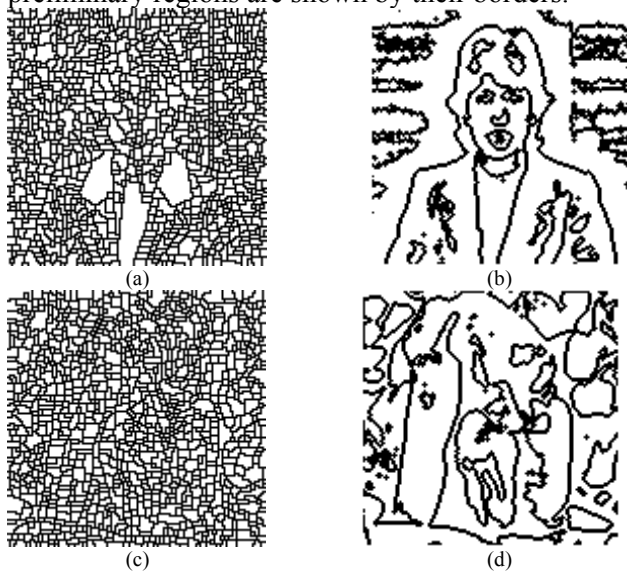

Fig. 20 The comparison between the preliminary regions obtained by the watershed algorithm and the proposed method
(a) and (c) are the results of the watershed algorithm; (b) and (d) are the results of the proposed method

From Fig. 20, it can be seen that the segmentation results of the two methods have quite different characteristics. The preliminary regions obtained by the proposed method are close to the ideal final segmentation results. However, those obtained by the watershed algorithm are relatively far from expected final segmentation results, although the border lines of the main objects can be identified by careful observation. Therefore, the watershed method surely needs more complex pre- or post- processing to reach a degree of practical use.

Experiments have also been done to compare the proposed method with the K-means method, which is widely and frequently used in data classification. Example results are shown in Fig. 21, where the segmentation results for Fig. 13(a) and Fig. 14(a) are shown respectively. The results indicate that the proposed method has better performance because the K-means method obtained relatively more cracked sub-regions within integral objects.

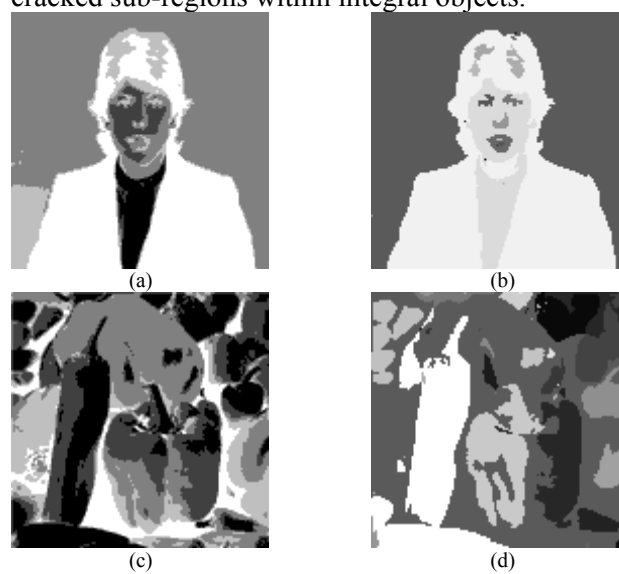

Fig. 21 The comparison between the segmentation results by the K-means segmentation method and the proposed method
(a) and (c) are the results of the K-means method; (b) and (d) are the results of the proposed method

## 4 Conclusion and Discussion

The dynamics of strain and stress is a basic and important topic in physics. The external force and the internal stress force caused by the object's deformation constitute the elements in a self-balancing mechanism. And self-balancing may be a required or even essential feature in solving many problems, which can make the method stable and produce meaningful and useful results.

In this paper, a model of virtual elastic mesh on the digital image is presented for region segmentation. In the model, the virtual repulsive force and the elastic force are defined by imitating the physical force but with slight difference to facilitate the segmentation problem. The repulsive force is defined according to greyscale difference between adjacent points, which is a "differentiation factor" representing the difference between adjacent regions. On the other hand, the elastic force is the "balancing factor", which homogenizes the height of the points within the same region where greyscale difference is small, at the same time counteracts the repulsive force at region borders where the greyscale difference is large. The final shape of the mesh at balance state is a representation of the image structure. And based on the sign distribution of the points' height, the region segmentation can be easily implemented by grouping or clustering the adjacent points of the same sign of height (i.e. *z* coordinate in the mesh).

The experimental results prove the effectiveness of the proposed method for several types of images, which is a preliminary but practical proof of the method's adaptability and robustness. On the other hand, the result comparison between the authors' previous work also proves the improvement brought by extending the model of virtual 2D grid to 3D mesh. For improvement of the method, detailed properties of the proposed model will be studied in future. In future study, experiments will be carried out to investigate the segmentation results under different set of parameters of the method, such as the constants defined for the simulation. And possible merging process which may be more reasonable will also be investigated as a necessary post-processing step to obtain better results of region merging.

*References:*
[1] Castrenze Polizzotto, A unifying variational framework for stress gradient and strain gradient elasticity theories, *European Journal of Mechanics, A/Solids*, Vol. 49, 2015, pp. 430-440.
[2] Robert Asaro, Vlado Lubarda, *Mechanics of Solids and Materials*, Cambridge University Press (Reissue edition), June 30, 2011.
[3] Castrenze Polizzotto, Stress gradient versus strain gradient constitutive models within elasticity, *International Journal of Solids and Structures*, Vol. 51, No. 9, 2014, pp. 1809-1818.
[4] William F. Hosford, *Solid Mechanics*, Cambridge University Press, March 22, 2010.
[5] Julia Burdge, *Chemistry*, McGraw-Hill Science, January 10, 2013.
[6] Amit Singh Chauhan, Sanjay Silakari, Manish Dixit, Image segmentation methods: A survey approach, *Proceedings of the 4th International Conference on Communication Systems and Network Technologies*, 2014, pp. 929-933.
[7] Yujin Zhang, *Advances in Image And Video Segmentation*, IRM Press, May 2, 2006.
[8] I.V. Gribkov, P.P. Koltsov, N.V. Kotovich, A.A. Kravchenko, A.S. Koutsaev, A.S. Osipov, A.V. Zakharov, Testing of image segmentation methods, *WSEAS Transactions on Signal Processing*, Vol. 4, No. 8, 2008, pp. 494-503.
[9] Lay Khoon Lee, Siau Chuin Liew, Weng Jie Thong, A review of image segmentation methodologies in medical image, *Lecture Notes in Electrical Engineering*, Vol. 315, 2015, pp. 1069-1080.
[10] N. Jain, A. Lala, Image segmentation: A short survey, *Confluence 2013: The Next Generation Information Technology Summit*, 2013, pp. 380-384.
[11] A. Gacsádi, L. Tepelea, I. Gavrilut, O. Straciuc, Energy based medical imaging segmentation methods by using Cellular Neural Networks, *Proceedings of the 15th WSEAS International Conference on Systems*, 2011, pp. 190-195.
[12] Xiaodong Zhuang, Nikos. Mastorakis, A Physics-Inspired Model of Image Structure Representation by Deformable Elastic Grid, *Proc. of the 7th WSEAS Int. Conf. on Computational Intelligence, Man-machine and Cybernetics*, pp. 209-214, 2008.
[13] Intan Aidha Yusoff, Nor Ashidi Mat Isa, Two-dimensional clustering algorithms for image segmentation, *WSEAS Transactions on Computers*, Vol. 10, No. 10, 2011, pp. 332-342.
[14] Mark S. Nixon, Xin U. Liu, Cem Direkoglu, David J. Hurley, On using physical analogies for feature and shape extraction in computer vision, *Computer Journal*, Vol. 54, No. 1, 2011, pp. 11-25.
[15] X. D. Zhuang, Nikos E. Mastorakis, The relative potential field as a novel physics-inspired method for image analysis, *WSEAS Transactions on Computers*, Vol. 9, No. 10, 2010, pp. 1086-1097.
[16] Mohammed M. Abdelsamea, Giorgio Gnecco, Mohamed Medhat Gaber, A Survey of SOM-Based Active Contour Models for Image Segmentation, *Advances in Intelligent Systems and Computing*, Vol. 295, 2014, pp. 293-302.
[17] X. D. Zhuang, N. E. Mastorakis, Image analysis based on the discrete magnetic field generated by the virtual edge current in digital images, *WSEAS Transactions on Computers*, Vol. 10, No. 8, 2011, pp. 259-273.
[18] Yuyu Liang, Mengjie Zhang, Will N. Browne, Image segmentation: A survey of methods based on evolutionary computation, *Lecture Notes in Computer Science*, Vol. 8886, 2014, pp. 847-859.